\documentclass[runningheads]{llncs}
\usepackage{graphicx}
\usepackage{amsmath}
\usepackage{amssymb}
\usepackage{booktabs}
\usepackage{times}
\usepackage{epsfig}
\usepackage{multirow}
\usepackage{bbding}
\begin{document}
\title{VCD: Visual Causality Discovery for Cross-Modal Question Reasoning}
%
%
\author{Yang Liu\inst{1}\orcidID{0000-0002-9423-9252} \and
Ying Tan\inst{1}\and
Jingzhou Luo\inst{1}\and
Weixing Chen\inst{1}}
\authorrunning{Yang Liu et al.}
%
\institute{Sun Yat-sen University, China\\
\email{liuy856@mail.sysu.edu.cn}\\
\email{tany86@mail2.sysu.edu.cn}\\
\email{luojzh5@mail2.sysu.edu.cn}\\
\email{chen867820261@gmail.com}}
\maketitle              
\vspace{-25pt}
\begin{abstract}
Existing visual question reasoning methods usually fail to explicitly discover the inherent causal mechanism and ignore jointly modeling cross-modal event temporality and causality. In this paper, we propose a visual question reasoning framework named Cross-Modal Question Reasoning (CMQR), to discover temporal causal structure and mitigate visual spurious correlation by causal intervention. To explicitly discover visual causal structure, the Visual Causality Discovery (VCD) architecture is proposed to find question-critical scene temporally and disentangle the visual spurious correlations by attention-based front-door causal intervention module named Local-Global Causal Attention Module (LGCAM). To align the fine-grained interactions between linguistic semantics and spatial-temporal representations, we build an Interactive Visual-Linguistic Transformer (IVLT) that builds the multi-modal co-occurrence interactions between visual and linguistic content. Extensive experiments on four datasets demonstrate the superiority of CMQR for discovering visual causal structures and achieving robust question reasoning.

\vspace{-5pt}
\end{abstract}

\vspace{-35pt}
\section{Introduction}
\vspace{-5pt}
Event understanding \cite{krishna2017dense,lei2020more} has become a prominent research topic in video analysis because videos have good potential to understand event temporality and causality. Since the expressivity of natural language can potentially describe a richer event space \cite{buch2022revisiting} that facilitates the deeper event understanding, in this paper, we focus on event-level visual question reasoning task, which aims to fully understand richer multi-modal event space and answer the given question in a causality-aware way. To achieve event-level visual question reasoning \cite{das2017visual,anderson2018vision}, the model is required to achieve fine-grained understanding of video and language content involving various complex relations such as spatial-temporal visual relation, linguistic semantic relation, and visual-linguistic causal dependency. Most of the existing visual question reasoning methods \cite{li2019beyond,le2020hierarchical} use recurrent neural networks (RNNs) \cite{sukhbaatar2015end}, attention mechanisms \cite{vaswani2017attention} or Graph Convolutional Networks \cite{kipf2016semi} for relation reasoning between visual and linguistic modalities. Although achieving promising results, the current visual question reasoning methods suffer from the following two common limitations.

\begin{figure}
\begin{center}
\includegraphics[scale=0.23]{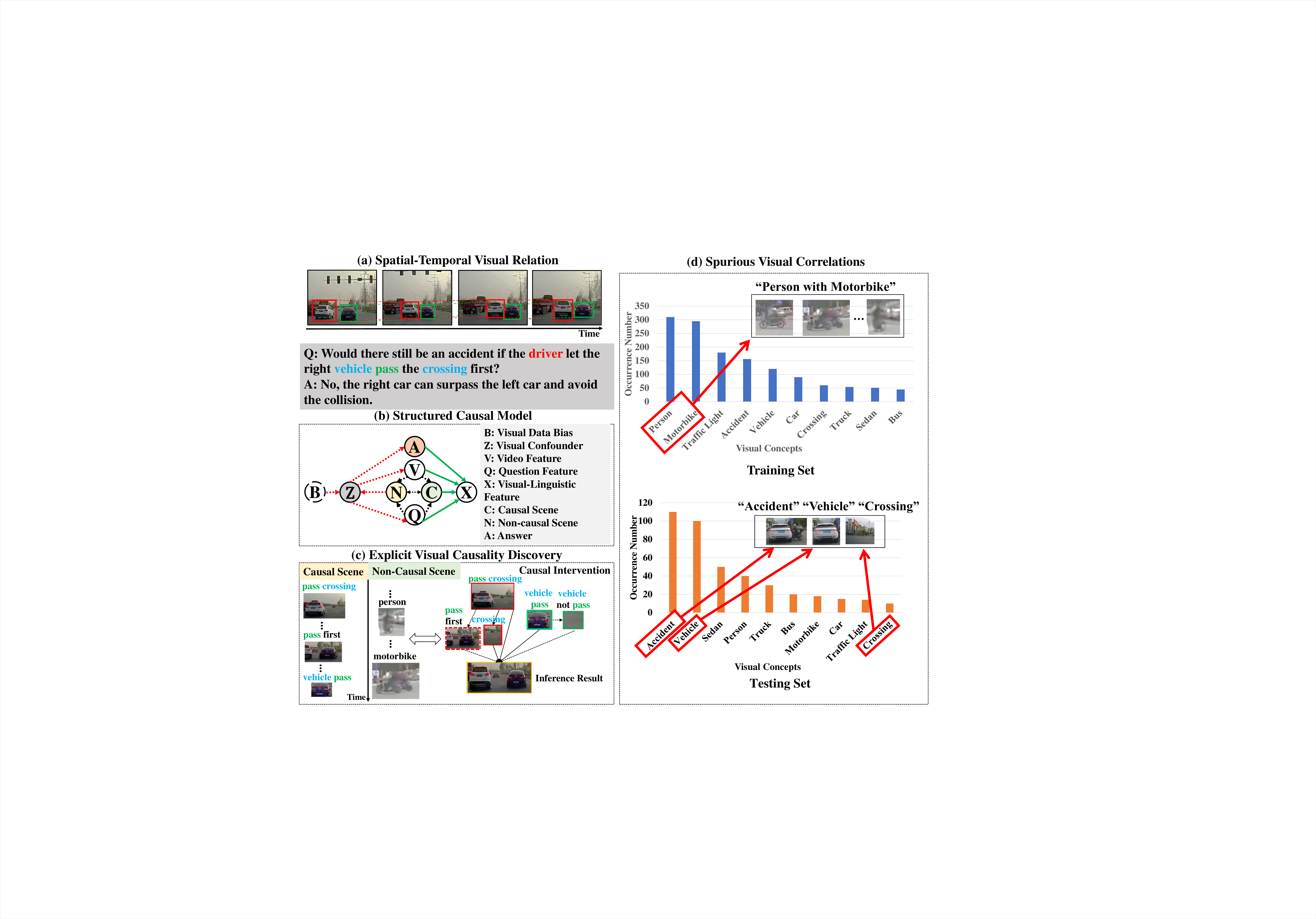}
\end{center}
    \vspace{-20pt}
   \caption{An example of event-level counterfactual visual question reasoning task and its structured causal model (SCM). The counterfactual inference is to obtain the outcome of certain hypothesis that do not occur. The SCM shows how the confounder induces the spurious correlation. The green path is the unbiased visual question reasoning. The red path is the biased one.}
      \vspace{-20pt}
\label{Fig1}
\end{figure}

First, the existing visual question reasoning methods usually focus on relatively simple events where temporal understanding and causality discovery are simply not required to perform well, and ignore more complex and challenging events that require in-depth understanding of the causality, spatial-temporal dynamics, and linguistic relations. As shown in Fig. \ref{Fig1} (a), the event-level counterfactual visual question reasoning task requires the outcome of certain hypothesis that does not occur in the given video. If we just simply correlate relevant visual contents, we cannot to get the right inference result without discovering the hidden spatial-temporal and causal dependencies. Moreover, the multi-level interaction and causal relations between the language and spatial-temporal structure is not fully explored in current methods.

Second, most of the visual question reasoning models tend to capture the spurious visual correlations rather than the true causal structure, which leads to an unreliable reasoning process \cite{niu2021counterfactual,wang2021causal,liu2022causalarxiv,liu2022causal,chen2023visual,liu2023causalvlr,10.1145/3581783.3611873,tang2023towards}. As shown in the SCM from Fig. \ref{Fig1} (b), we can consider some frequently appearing visual concepts as the visual confounders. The ``visual  bias'' denotes the strong correlations between visual features and answers. For example, the concepts ``person" and ``motorbike'' are dominant in training set (Fig. \ref{Fig1} (c) and (d)) and thus the predictor may learn the spurious correlation between the ``person" with the ``motorbike'' without looking at the collision region (causal scene $C$) to reason how actually the accident happens. Taking a causal look at VideoQA, we partition the visual scenes into two parts: 1) causal scene $C$, which holds the question-critical information, 2) non-causal scene $N$, which is irrelevant to the answer. Such biased dataset entails two causal effects: the visual bias $B$ and non-causal scene $N$ lead to the confounder $Z$, and then affects the visual feature $V$, causal scene $C$, question feature $Q$, visual-linguistic feature $X$, and the answer $A$. As shown in Fig. \ref{Fig1} (d),  due to the existence of  visual confounders and  visual-linguistic interaction, the model learns the spurious correlation without exploiting the true question intention and dominant visual evidence.

To address the aforementioned limitations, we propose an event-level visual question reasoning framework named  Cross-Modal Question Reasoning  (CMQR). Experiments on SUTD-TrafficQA, TGIF-QA, MSVD-QA, and MSRVTT-QA datasets show the advantages of our CMQR over the state-of-the-art methods. The main contributions can be summarized as follows:
\vspace{-8pt}
\begin{itemize}
\setlength{\itemsep}{0pt}
\setlength{\parsep}{0pt}
\setlength{\parskip}{0pt}
\item We propose a causality-aware visual question reasoning framework named Cross-Modal Question Reasoning (CMQR), to discover cross-modal causal structures via causal interventions and achieve robust visual question reasoning and answering.
\item We introduce the Visual Causality Discovery (VCD) architecture that learns to find the temporal causal scenes for a given question and mitigates the unobservable visual spurious correlations by an attention-based causal front-door intervention module named Local-Global Causal Attention Module (LGCAM).
\item We construct an Interactive Visual-Linguistic Transformer (IVLT) to align and discover the multi-modal co-occurrence interactions between linguistic semantics and spatial-temporal visual concepts.
\end{itemize}

\vspace{-20pt}
\section{Related Works}
\vspace{-5pt}
\subsection{Visual Question Reasoning}
\vspace{-5pt}
Compared with the image-based visual question reasoning \cite{antol2015vqa,yang2016stacked,anderson2018bottom}, event-level visual question reasoning is much more challenging due to the existence of temporal dimension. To accomplish this task, the model needs to capture spatial-temporal \cite{liu2016combining,liu2018global,liu2018transferable,lan2022audio,zhu2022hybrid,wang2023urban,lin2023denselight,liu2023self,yan2023skeletonmae} and visual-linguistic relations. To explore relational reasoning, Xu et al. \cite{xu2017video} proposed an attention mechanism to exploit the appearance and motion knowledge with the question as a guidance. Later on, some hierarchical attention and co-attention based methods \cite{fan2019heterogeneous,jiayincai2020feature,le2020hierarchical,huang2020location,lei2021less,liu2021hair} are proposed to learn appearance-motion and question-related multi-modal interactions.  Le et al. \cite{le2020hierarchical} proposed hierarchical conditional relation network (HCRN) to construct sophisticated structures for representation and reasoning over videos. Lei et al. \cite{lei2021less} employed sparse sampling to build a transformer-based model named CLIPBERT and achieve video-and-language understanding. However, previous works tend to implicitly capture the spurious visual-linguistic correlations, while we propose the Visual Causality Discovery (VCD) to explicitly uncover the visual causal structure.

\vspace{-15pt}
\subsection{Causal Inference in Visual Learning}
\vspace{-5pt}
Compared with conventional debiasing techniques \cite{wang2020devil}, causal inference \cite{pearl2016causal,yang2021deconfounded,liu2022causal} shows its potential in mitigating the spurious correlations \cite{bareinboim2012controlling} and disentangling the desired model effects \cite{besserve2020counterfactuals} for better generalization. Counterfactual and causal inference have attracted increasing attention in visual explanations \cite{goyal2019counterfactual,hendricks2018grounding,wang2020scout}, scene graph generation \cite{chen2019counterfactual,tang2020unbiased}, image recognition \cite{wang2020visual,wang2021causal}, video analysis \cite{fang2019modularized,liu2018hierarchically,kanehira2019multimodal,liu2019deep,liu2021semantics,ni2022cross,liu2022tcgl,nan2021interventional}, and vision-language tasks \cite{cao2020strong,le2019choi,niu2021counterfactual,yang2021causal,li2022invariant,liu2022cross,liu2023causality,CMCIR,chen2023visual,wei2023visual}. However, most of the existing visual tasks are relatively simple. Although some recent works CVL \cite{abbasnejad2020counterfactual}, Counterfactual VQA \cite{niu2021counterfactual}, CATT \cite{yang2021causal}, and IGV \cite{li2022invariant} focused on visual question reasoning tasks, they adopted structured causal model (SCM) to eliminate either the linguistic or visual bias without considering explicit cross-modal causality discovery. Differently,  our CMQR aims for event-level visual question reasoning that requires fine-grained understanding of spatial-temporal and visual-linguistic causal dependency. Moreover, our Visual Causality Discovery (VCD) applies front-door causal interventions to explicitly find question-critic visual scene .

\vspace{-10pt}
\section{Methodology}
\vspace{-10pt}
The CMQR is an event-level visual question reasoning architecture, as shown in Fig. \ref{Fig2}. In this section, we present the detailed implementations of CMQR.

\vspace{-10pt}
\subsection{Visual Representation Learning}
\vspace{-5pt}
The goal of event-level visual question reasoning is to deduce an answer $\tilde{a}$ from a video $\mathcal{V}$ with a given question $q$. The video $\mathcal{V}$ of $L$ frames is divided into $N$ equal clips. Each clip of $C_i$ of length $T=\lfloor L/N\rfloor$ is presented by two types of visual features: frame-wise appearance feature vectors $F^a_i=\{f_{i,j}^a|f_{i,j}^a,  j=1,\ldots,T\}$ and motion feature vector at clip level $f^m_i$. In our experiments, the vision-language transformer with frozen parameters XCLIP \cite{ni2022expanding} (other visual backbones are evaluated in Table \ref{Table11}) is used to extract the frame-level appearance features $F^a$ and the clip-level motion features $F^m$. We use a linear layer to map $F^a$ and $F^m$ into the same $d$-dimensional feature space.

\vspace{-10pt}
\subsection{Linguistic Representation Learning}\label{sec3.3}
\vspace{-5pt}
Each word of the question is respectively embedded into a vector of $300$ dimension by Glove \cite{pennington2014glove} word embedding, which is further mapped into a $d$-dimensional space using linear transformation. Then, we represent the corresponding question and answer semantics as $Q=\{q_1,q_2,\cdots, q_L\}$, $A=\{a_1,a_2,\cdots, a_{L_a}\}$, where $L$, $L_a$ indicate the length of $Q$ and $A$, respectively. To obtain contextual linguistic representations that aggregates dynamic long-range temporal dependencies from multiple time-steps, a Bert \cite{devlin2018bert} model is employed to encode $Q$ and the answer $A$, respectively. Finally, the updated representations for the question and answer candidates can be written as:
\vspace{-5pt}
\begin{equation}\label{eq2}
\begin{aligned}
&Q=\{q_i|q_i\in \mathbb{R}^{d}\}_{i=1}^{L}\\
&A=\{a_i|a_i\in \mathbb{R}^{d}\}_{i=1}^{L_a}
\end{aligned}
\end{equation}

\begin{figure*}[t]
\begin{center}
\includegraphics[scale=0.175]{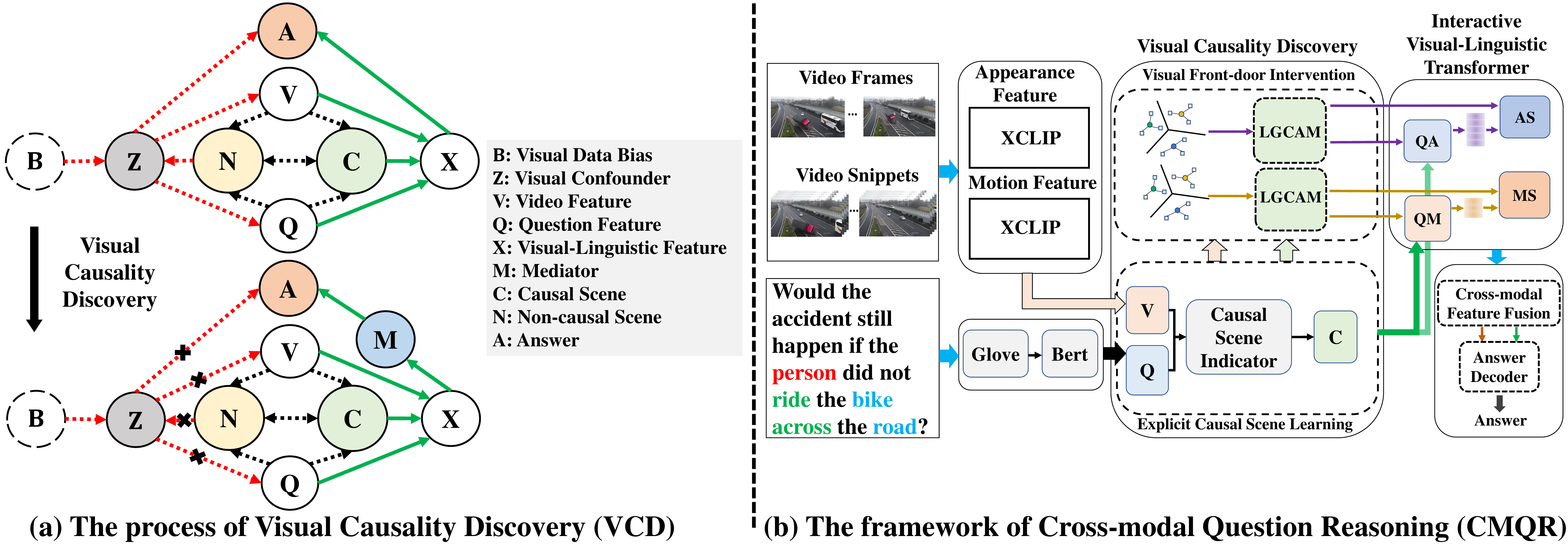}
\end{center}
   \vspace{-18pt}
   \caption{The process of Visual Causality Discovery  (VCD)  ((a)) and framework of Cross-modal Question Reasoning ((b)). The green path is the unbiased question reasoning. The red path is the biased one. The black path is the division of causal and non-causal visual scenes. }
      \vspace{-15pt}
\label{Fig2}
\end{figure*}

   \vspace{-15pt}
\subsection{Visual Causality Discovery}
   \vspace{-5pt}
For visual-linguistic question reasoning, we employ Pearl's structural causal model (SCM) \cite{pearl2016causal} to model the causal effect between video-question pairs $(V,Q)$, causal scene $C$, non-causal scene $N$, and the answer $A$, as shown in Fig. \ref{Fig2} (a). We hope to train a video question answering model to the learn the true causal effect $\{V,Q\}\rightarrow C\rightarrow X\rightarrow A$: the model should reason the answer $A$ from video feature $V$, causal-scene $C$ and question feature $Q$ instead of exploiting the non-causal scene $N$ and spurious correlations induced by the confounders $Z$ (i.e., the existence of non-causal scene and overexploiting the co-occurrence between visual concepts and answer). In our SCM, the non-interventional prediction can be expressed by Bayes rule:
\vspace{-5pt}
\begin{equation}\label{eq4}
P(A|V,Q)=\sum_zP(A|V,Q,z)P(z|V,Q)
\end{equation}
However, the above objective learns not only the main direct correlation from $\{V,Q\}\rightarrow X\rightarrow A$ but also the spurious one from the back-door path $\{V,Q\}\leftarrow Z\rightarrow A$. An intervention on $\{V,Q\}$ is denoted as $do(V,Q)$, which cuts off the link $\{V,Q\}\leftarrow Z$ to block the back-door path $\{V,Q\}\leftarrow Z\rightarrow A$ and eliminate the spurious correlation. In this way, $\{V,Q\}$ and $A$ are deconfounded and the model can learn the true causal effect $\{V,Q\}\rightarrow C\rightarrow X\rightarrow A$.

\vspace{-10pt}
\subsubsection{Explicit Causal Scene Learning}
Inspired by the fact that only part of the visual scenes are critical to answering the question, we split the video $V$ into causal scene $C$ and non-causal scene $N$ (see the black path in Fig. \ref{Fig2} (a)). Specifically, given the causal scene $C$ and question $Q$, we assume that the answer $A$ is determined, regardless the variations of the non-causal scene N: $A\perp N|C,Q$, where $\perp$ denotes the probabilistic independence. Thus, we build an explicit causal scene learning (ECSL) module to estimate $C$.

For a video-question pair $(v,q)$, we encode video instance $v$ as a sequence of $K$ visual clips. The ECSL module aims to estimate the causal scene $\hat{c}$ according to the question $q$. Concretely, we first construct a cross-modal attention module to indicate the probability of each visual clip belongs to causal scene ($p_{\hat{c}}\in \mathbb{R}^K$):
\vspace{-5pt}
\begin{equation}\label{eq5}
\begin{aligned}
p_{\hat{c}}=\textrm{softmax}(G_v^1(v)\cdot G_q^1(q)^\top)\\
\end{aligned}
\end{equation}
where $G_v^1$ and $G_q^1$  are fully connected layers to align cross-modal representations. However, the soft mask makes $\hat{c}$ overlap. To achieve a differentiable selection on attentive probabilities and compute the selector vector $S\in \mathbb{R}^{K}$ on the attention score over each clip (i.e., $p_{\hat{c},i}, i\in K$), we employ Gumbel-Softmax \cite{jang2016categorical} and estimate $\hat{c}$ as:
\vspace{-5pt}
\begin{equation}\label{eq6}
\hat{c}=\textrm{Gumbel-Softmax}(p_{\hat{c},i})\cdot v
\end{equation}
\vspace{-15pt}

For a video-question pair $(v,q)$, we obtain the original video $v$ and causal scene $\hat{c}$. According to Eq. (\ref{eq4}), we pair original video $v$ and causal scene $\hat{c}$ with $q$ to synthesizes two instances: $(v,q)$ and $(\hat{c},q)$. Then, we feed these two instances into visual front-door causal intervention (VFCI) module to deconfound $\{V,Q\}$ and $A$.

\vspace{-10pt}
\subsubsection{Visual Front-door Causal Intervention}
In visual domains, it is hard to explicitly represent confounders due to complex data biases. Fortunately, the front-door adjustment give a feasible way to calculate $P(A|do(V),Q)$. In Fig. \ref{Fig2} (a), an additional mediator $M$ can be inserted between $X$ and $A$ to construct the front-door path $\{V,Q\}\rightarrow X\rightarrow M\rightarrow A$ for transmitting knowledge. For visual question reasoning, an attention-based model $P(A|V,Q)=\sum_mP(M=m|V,Q)P(A|M=m)$ will select a few regions from the original video $V$ and causal scene $C$ based on the question $Q$ to predict the answer $A$, where $m$ denotes the selected knowledge from $M$. Thus, the answer predictor can be represented by two parts: two feature extractors $V\rightarrow X\rightarrow M$, $C\rightarrow X\rightarrow M$,  and an answer predictor $M\rightarrow A$. In the following, we take the visual interventional probability $P(A|do(V),Q)$ for original video $V$ as an example (the $P(A|do(C),Q)$ for causal scene $C$ is implemented in the same way):
\vspace{-5pt}
\begin{equation}\label{eq11}
\begin{aligned}
&P(A|do(V),Q)=\\
&=\sum_mP(M=m|do(V),Q)P(A|do(M=m))\\
&=\sum_mP(M=m|V,Q)\sum_vP(V=v)P(A|V=v,M=m)
\end{aligned}
\end{equation}

To implement visual front-door causal intervention Eq. (\ref{eq11}) in a deep learning framework, we parameterize the $P(A|V,M)$ as a network $g(\cdot)$ followed by a softmax layer since most of visual-linguistic tasks are classification formulations, and then apply Normalized Weighted Geometric Mean (NWGM) \cite{xu2015show} to reduce computational cost:
\vspace{-5pt}
\begin{equation}\label{eq13}
\begin{aligned}
&P(A|do(V),Q)\approx \textrm{Softmax}[g(\hat{M},\hat{V})]\\
&=\textrm{Softmax}\Big[g(\sum_mP(M=m|f(V))m,\sum_vP(V=v|h(V))v)\Big]
\end{aligned}
\end{equation}
where $\hat{M}$ and $\hat{V}$ denote the estimations of $M$ and $V$, $h(\cdot)$ and $f(\cdot)$ denote the network mappings. The derivation details from Eq. (\ref{eq11})-(\ref{eq13}) is given in the Appendix 2.

Actually, $\hat{M}$ is essentially an in-sample sampling process where $m$ denotes the selected knowledge from the current input sample $V$, $\hat{V}$ is essentially  a cross-sample sampling process since it comes from the other samples. Therefore, both $\hat{M}$ and $\hat{V}$ can be calculated by attention networks \cite{yang2021causal}.


Therefore, we propose a Local-Global Causal Attention Module (LGCAM) that jointly estimates $\hat{M}$ and $\hat{V}$ to increase the representation ability of the causality-aware visual features. $\hat{M}$ can be learned by local-local visual feature $F_{LL}$, $\hat{V}$ can be learned by local-global visual feature $F_{LG}$. Here, we take the computation of $F_{LG}$ as the example to clarify our LGCAM. Specifically, we firstly calculate $F_L=f(V)$ and $F_G=h(V)$ and use them as the input, where $f(\cdot)$ denotes the visual feature extractor (frame-wise appearance feature or motion feature) followed by a query embedding function, and $h(\cdot)$ denotes the K-means based visual feature selector from the whole training samples followed by a query embedding function. Thus, $F_L$ represents the visual feature of the current input sample (local visual feature) and $F_G$ represents the global visual feature. The $F_G$ is obtained by randomly sampling from the whole clustering dictionaries with the same size as $F_L$. The LGCAM takes $F_L$ and $F_G$ as the inputs and computes local-global visual feature $F_{LG}$ by conditioning global visual feature $F_G$ to the local visual feature $F_L$. The output of the LGCAM is denoted as $F_{LG}$:
\vspace{-8pt}
\begin{equation}\label{eq14}
\begin{aligned}
&\textbf{\textrm{Input}}: Q=F_L, K=F_G, V=F_G\\
&\textbf{\textrm{Local-Global}}: H=[W_VV,W_QQ\odot W_KK]\\
&\textbf{\textrm{Activation Mapping}}: H^\prime=\textrm{GELU}(W_HH+b_H)\\
&\textbf{\textrm{Attention Weights}}: \alpha=\textrm{Softmax}(W_{H^\prime}H^\prime+b_{H^\prime})\\
&\textbf{\textrm{Output}}: F_{LG}=\alpha\odot F_G
\end{aligned}
\end{equation}
where $[.,.]$ denotes concatenation operation, $\odot$ is the Hadamard product, $W_Q$, $W_K$, $W_V$, $W_{H^\prime}$ denote the weights of linear layers,  $b_H$ and $b_{H^\prime}$  denote the biases of linear layers. From Fig. \ref{Fig2} (b), the visual front-door causal intervention module has two branches for appearance and motion features. Therefore, the  $F_{LG}$ has two variants, one for appearance branch $F_{LG}^a$, and the other for motion branch $F_{LG}^m$. The $F_{LL}$ can be computed similarly as $F_{LG}$ when setting $Q=K=V=F_L$. Finally, the $F_{LG}$ and $F_{LL}$ are concatenated $F_C=[F_{LG},F_{LL}]$ for estimating $P(A|do(V),Q)$.


\vspace{-10pt}
\subsection{Interactive Visual-Linguistic Transformer}
\vspace{-6pt}
To align the fine-grained interactions between linguistic semantics and spatial-temporal representations, we build an Interactive Visual-Linguistic Transformer (IVLT) that contains four sub-modules, namely Question-Appearance (QA), Question-Motion (QM), Appearance-Semantics (AS) and Motion-Semantics (MS).  The QA (QM) module consists of an R-layer Multi-modal Transformer Block (MTB)  for multi-modal interaction between the question and the appearance (motion) features. Similarly, the AS (MS) uses the MTB to infer the appearance (motion) information given the questions.

For QA and QM modules, the input of MTB is question representation $Q$ obtained from section 3.2 and causality-aware visual representations $F_C^a$,  $F_C^m$ obtained from section 3.3.2,  respectively. To maintain the positional information of the video sequence, the appearance feature $F_C^a$ and motion feature $F_C^m$ are firstly added with the learned positional embeddings $P^a$ and $P^m$, respectively. Thus, for $r=1,2,\ldots, R$ layers of the MTB, with the input $F_C^a=[F_C^a,P^a]$, $F_C^m=[F_C^m,P^m]$, $Q^a$, and $Q^m$, the multi-modal output for QA and QM are computed as:
\vspace{-6pt}
\begin{equation}\label{eq15}
\begin{aligned}
&\hat{Q}^a_r=U^a_r+\sigma^a(\textrm{LN}(U^a_r))\\
&\hat{Q}^m_r=U^m_r+\sigma^m(\textrm{LN}(U^m_r))\\
&U^a_r=\textrm{LN}(\hat{Q}^a_{r-1})+\textrm{MMA}^a(\hat{Q}^a_{r-1},F_C^a)\\
&U^m_r=\textrm{LN}(\hat{Q}^m_{r-1})+\textrm{MMA}^m(\hat{Q}^m_{r-1},F_C^m)\\
\end{aligned}
\end{equation}
where $\hat{Q}^a_0=Q^h, \hat{Q}^m_0=Q^h$, $U^a_r$ and $U^m_r$ are the intermediate feature at $r$-th layer of the MTB.  LN$(\cdot)$ denotes the layer normalization operation and $\sigma^a(\cdot)$  and $\sigma^m(\cdot)$ denote the two-layer linear projections with GELU activation. MMA$(\cdot)$ is the Multi-head Multi-modal Attention layer. We denote the output semantics-aware appearance and motion features of QA and MA as $L^a=\hat{Q}^a=\hat{Q}^a_R$ and $L^m=\hat{Q}^m=\hat{Q}^m_R$, respectively.

Similar to Eq. (\ref{eq15}), given the visual appearance and motion feature ${F}_{LG}^a$, ${F}_{LG}^m$ and question semantics $L^a$, $L^m$, the multi-modal output for AS and MS are computed as:
\vspace{-5pt}
\begin{equation}\label{eq18}
\begin{aligned}
&\hat{L}^a_r=U^a_r+\sigma^a(\textrm{LN}(U^a_r))\\
&\hat{L}^m_r=U^m_r+\sigma^m(\textrm{LN}(U^m_r))\\
&U^a_r=\textrm{LN}(F_{C, r-1}^a)+\textrm{MMA}^a(F_{C, r-1}^a, L^a)\\
&U^m_r=\textrm{LN}(F_{C, r-1}^m)+\textrm{MMA}^m(F_{C, r-1}^m, L^m)\\
\end{aligned}
\end{equation}
where $F_{C, 0}^a=F_{C}^a$, $F_{C, 0}^m=F_{C}^m$. The output visual clues of QA and MA are denoted as $F_{s}^a=\hat{L}^a_R$ and $F_{s}^m=\hat{L}^m_R$, respectively. Then, the output of the AS and MS are concatenated to make the final visual output $F=[F_{s}^a,F_{s}^m]\in \mathbb{R}^{2d}$. The output of the QA and QM are concatenated to make the  final question semantics output $L=[L^a,L^m]\in \mathbb{R}^{2d}$.

\vspace{-10pt}
\subsection{Cross-modal Feature Fusion and Training}
\vspace{-5pt}
For $(v,q)$ and $(\hat{c},q)$, their visual and linguistic outputs of the IVLT model are denoted as $F, F_c$ and $L, L_c$, respectively. Inspired by the adaptive feature fusion \cite{liu2022tcgl} (refer to Appendix 1), we obtain the refined linguistic feature vectors $\{\widetilde{L}, \widetilde{L}_c\}$, which are then concatenated to form the final semantic-ware linguistic feature $\widetilde{L}=[\widetilde{L}, \widetilde{L}_c]\in \mathbb{R}^{2d}$.

To obtain the semantic-aware visual feature, we compute the visual feature $\widetilde{F}_{k}$ by individually conditioning each instance from visual features $\{F_1, F_2\}=\{F, F_c\}$ to each instance from refined linguistic features $\{\widetilde{L}_1, \widetilde{L}_2\}=\{\widetilde{L}, \widetilde{L}_c\}$ using the same operation as \cite{le2020hierarchical}. Then, these semantic-aware visual features $\widetilde{F}_{k}~(k=1,2)$ are concatenated to form the final semantic-aware visual feature $\widetilde{F} \in \mathbb{R}^{2d}$.

Finally, we apply different answer decoders \cite{le2020hierarchical} to $(v,q)$ and $(\hat{c},q)$ and obtain original prediction and causal prediction losses:
\vspace{-10pt}
\begin{equation}\label{eq19}
\begin{aligned}
&\mathcal{L}_{o}=\textrm{XE}(\textrm{CMQR}(v, q),a)\\
&\mathcal{L}_{c}=\textrm{XE}(\textrm{CMQR}(\hat{c}, q),a)
\end{aligned}
\end{equation}
where XE denotes the cross-entropy loss, $a$ is the ground-truth answer, CMQR denotes our proposed framework. Furthermore, to make the predictions of original and causal scene consistent, we apply KL-divergence between the predictions of $(v,q)$ and  $(\hat{c},q)$:
\begin{equation}\label{eq20}
\mathcal{L}_{a}=\textrm{KL}(\textrm{CMQR}(v, q),\textrm{CMQR}(\hat{c}, q))
\end{equation}

Finally, the learning objective of our CMQR is:
\begin{equation}\label{eq20}
\mathcal{L}_{\textrm{CMQR}}=\mathcal{L}_{o}+\lambda_c\mathcal{L}_{c}+\lambda_a\mathcal{L}_{a}
\end{equation}

\vspace{-15pt}
\section{Experiments}
\vspace{-10pt}
\subsection{Datasets}
\vspace{-5pt}
In this paper, we evaluate our CMQR on four VideoQA datasets. \textbf{SUTD-TrafficQA} \cite{xu2021sutd} consists of 62,535 QA pairs and 10,090 traffic videos. There are six challenging reasoning tasks including basic understanding, event forecasting, reverse reasoning, counterfactual inference, introspection and attribution analysis. \textbf{TGIF-QA} \cite{jang2017tgif} has 165K QA pairs collected from 72K animated GIFs. It has four tasks: repetition count, repeating action, state transition, and frame QA. \textbf{MSVD-QA} \cite{xu2017video} contains 50,505 algorithm-generated question-answer pairs and 1,970 trimmed video clips. \textbf{MSRVTT-QA} \cite{xu2017video} contains 10,000 trimmed video clips and 243,680 question-answer pairs. More details of these datasets are given in Appendix 4.
\vspace{-15pt}
\subsection{Implementation Details}
\vspace{-5pt}
For fair comparisons, we follow \cite{le2020hierarchical} to divide the videos into 8 clips for all datasets. The XCLIP \cite{liu2021swin} with ViT-L/14 pretrained on Kinetics-600 dataset is used to extract the appearance and motion features. For the question, we adopt the pre-trained $300$-dimensional Glove \cite{pennington2014glove} word embeddings to initialize the word features in the sentence. For parameter settings, we set the dimension $d$ of hidden layer to 512. For the Multi-modal Transformer Block (MTB), the number of layers $r$ is set to 3 for SUTD-TrafficQA, $8$ for TGIF-QA,  $5$ for MSVD-QA, and $6$ for MSRVTT-QA. The number of attentional heads $H$ is set to 8. The dictionary is initialized by applying K-means over the whole visual features from the whole training set to get $512$ clusters and is updated during end-to-end training. We train the model using the Adam optimizer with an initial learning rate 2e-4, a momentum 0.9, and a weight decay 0. The learning rate reduces by half when the loss stops decreasing after every  $5$ epochs. The batch size is set to 64. All experiments are terminated after $50$ epochs. $\lambda_c$ and $\lambda_a$ are all set to $0.1$.

\begin{table}[t]\renewcommand\tabcolsep{3pt}\renewcommand\arraystretch{0.9}\scriptsize
\begin{center}
\begin{tabular}{lccccccc}
\hline
Method&Basic&Attri.&Intro.&Counter.&Fore.&Rev.&All\\\hline
$\textrm{VQAC}^\dag$  \cite{kim2021video}&34.02&49.43&\underline{34.44}&39.74&38.55&49.73&36.00\\
$\textrm{MASN}^\dag$ \cite{seo2021attend}&33.83&\underline{50.86}&34.23&41.06&41.57&\underline{50.80}&36.03\\
$\textrm{DualVGR}^\dag$ \cite{wang2021dualvgr}&33.91&50.57&33.40&\underline{41.39}&41.57&50.62&36.07\\
HCRN \cite{le2020hierarchical}&-&-&-&-&-&-&36.49\\
$\textrm{HCRN}^\dag$ \cite{le2020hierarchical}&\underline{34.17}&50.29&33.40&40.73&\underline{44.58}&50.09&36.26\\
Eclipse \cite{xu2021sutd}&-&-&-&-&-&-&37.05\\
$\textrm{IGV}^\dag$ \cite{li2022invariant}&-&-&-&-&-&-&\underline{37.71}\\\hline
\textbf{CMQR (ours)}&\textbf{36.10}&\textbf{52.59}&\textbf{38.38}&\textbf{46.03}&\textbf{48.80 }&\textbf{58.05}&\textbf{38.63}\\\hline
\end{tabular}
\end{center}
\vspace{-10pt}
\caption{Results on SUTD-TrafficQA dataset. }
\vspace{-20pt}
\label{Table4}
\end{table}

\begin{table}[t]\renewcommand\tabcolsep{6pt}\renewcommand\arraystretch{0.95}\scriptsize
\begin{center}
\begin{tabular}{lcccc}
\hline
Method&Action$\uparrow$&Transition$\uparrow$&FrameQA$\uparrow$&Count$\downarrow$\\\hline
ST-VQA  \cite{jang2017tgif}&62.9&69.4&49.5&4.32\\
Co-Mem  \cite{gao2018motion}&68.2&74.3&51.5&4.10\\
PSAC  \cite{li2019beyond}&70.4&76.9&55.7&4.27\\
HME \cite{fan2019heterogeneous}&73.9&77.8&53.8&4.02\\
GMIN \cite{gu2021graph}&73.0&81.7&57.5&4.16\\
L-GCN \cite{huang2020location}&74.3&81.1&56.3&3.95\\
HCRN \cite{le2020hierarchical}&75.0&81.4&55.9&3.82\\
HGA \cite{jiang2020reasoning}&75.4&81.0&55.1&4.09\\
QueST \cite{jiang2020divide}&75.9&81.0&59.7&4.19\\
Bridge \cite{park2021bridge}&75.9&\underline{82.6}&57.5&\textbf{3.71}\\
QESAL\cite{liu2021question}&76.1&82.0&57.8&3.95\\
ASTG \cite{jin2021adaptive}&76.3&82.1&\underline{61.2}&\underline{3.78}\\
CASSG \cite{liu2022cross}&77.6&\textbf{83.7}&58.7&3.83\\
HAIR \cite{liu2021hair}&77.8&82.3&60.2&3.88\\\hline
\textbf{CMQR (ours) }&\textbf{78.1}&82.4&\textbf{62.3}&3.83\\\hline
\end{tabular}
\end{center}
\vspace{-10pt}
\caption{Comparison with state-of-the-art methods on TGIF-QA. }
\vspace{-30pt}
\label{Table5}
\end{table}

\begin{minipage}{\textwidth}\scriptsize
\begin{minipage}[t]{0.49\textwidth}
\makeatletter\def\@captype{table}
\begin{tabular}{lcccccc}
\hline
Method&What &Who&How&When&Where&All\\\hline
HGA  \cite{jiang2020reasoning}&23.5&50.4&83.0&72.4&46.4&34.7\\
GMIN \cite{gu2021graph}&24.8&49.9&84.1&\underline{75.9}&\textbf{53.6}&35.4\\
QueST \cite{jiang2020divide}&24.5&52.9&79.1&72.4&\underline{50.0}&36.1\\
HCRN \cite{le2020hierarchical}&-&-&-&-&-&36.1\\
CASSG \cite{liu2022cross}&24.9&52.7&\textbf{84.4}&74.1&\textbf{53.6}&36.5\\
QESAL\cite{liu2021question}&25.8&51.7&83.0&72.4&\underline{50.0}&36.6\\
Bridge \cite{park2021bridge}&-&-&-&-&-&37.2\\
HAIR \cite{liu2021hair}&-&-&-&-&-&37.5\\
VQAC \cite{kim2021video}&26.9&53.6&-&-&-&37.8\\
MASN \cite{seo2021attend}&-&-&-&-&-&38.0\\
HRNAT \cite{gao2022hierarchical}&-&-&-&-&-&38.2\\
ASTG \cite{jin2021adaptive}&26.3&\underline{55.3}&82.4&72.4&\underline{50.0}&38.2\\
DualVGR \cite{wang2021dualvgr}&\underline{28.6}&53.8&80.0&70.6&46.4&39.0\\
IGV \cite{li2022invariant}&-&-&-&-&-&\underline{40.8}\\
\textbf{CMQR (Ours)}&\textbf{37.0}&\textbf{59.9}&\underline{81.0}&\textbf{75.8}&46.4&\textbf{46.4}\\\hline
\end{tabular}
\vspace{-10pt}
\caption{Comparison with state-of-the-art methods on MSVD-QA.  }
\label{Table6}
\end{minipage}
\begin{minipage}[t]{0.49\textwidth}\scriptsize
\makeatletter\def\@captype{table}
\begin{tabular}{lcccccc}
\hline
Method&What &Who&How&When&Where&All\\\hline
QueST \cite{jiang2020divide}&27.9&45.6&83.0&75.7&31.6&34.6\\
HRA \cite{chowdhury2018hierarchical}&-&-&-&-&-&35.0\\
MASN \cite{seo2021attend}&-&-&-&-&-&35.2\\
HRNAT \cite{gao2022hierarchical}&-&-&-&-&-&35.3\\
HGA \cite{jiang2020reasoning}&29.2&45.7&83.5&75.2&34.0&35.5\\
DualVGR \cite{wang2021dualvgr}&29.4&45.5&79.7&76.6&36.4&35.5\\
HCRN \cite{le2020hierarchical}&-&-&-&-&-&35.6\\
VQAC \cite{kim2021video}&29.1&46.5&-&-&-&35.7\\
CASSG \cite{liu2022cross}&29.8&46.3&\textbf{84.9}&75.2&35.6&36.1\\
GMIN \cite{gu2021graph}&30.2&45.4&\underline{84.1}&74.9&\textbf{43.2}&36.1\\
QESAL\cite{liu2021question}&30.7&46.0&82.4&76.1&\underline{41.6}&36.7\\
Bridge \cite{park2021bridge}&-&-&-&-&-&36.9\\
HAIR \cite{liu2021hair}&-&-&-&-&-&36.9\\
ClipBert \cite{lei2021less}&-&-&-&-&-&37.4\\
ASTG \cite{jin2021adaptive}&\underline{31.1}&\underline{48.5}&83.1&\underline{77.7}&38.0&37.6\\
IGV \cite{li2022invariant}&-&-&-&-&-&\underline{38.3}\\
\textbf{CMQR (ours)}&\textbf{32.2}&\textbf{50.2}&82.3&\textbf{78.4}&38.0&\textbf{38.9}\\\hline
\end{tabular}
\vspace{-10pt}
\caption{Comparison with state-of-the-art methods on MSRVTT. }
\vspace{5pt}
\label{Table7}
\end{minipage}
\end{minipage}
\vspace{-5pt}
\vspace{-10pt}
\subsection{Comparison with State-of-the-art Methods}
\vspace{-5pt}
The results in Table \ref{Table4} demonstrate that our CMQR achieves the best performance for six reasoning tasks including basic understanding, event forecasting, reverse reasoning, counterfactual inference, introspection and attribution analysis. Specifically, the CMQR improves the state-of-the-art methods Eclipse \cite{xu2021sutd} and IGV \cite{li2022invariant} by $1.58\%$ and $0.92\%$. Compared with the re-implemented methods $\textrm{VQAC}^\dag$, $\textrm{MASN}^\dag$, $\textrm{DualVGR}^\dag$, $\textrm{HCRN}^\dag$ and $\textrm{IGV}^\dag$, our CMQR outperforms these methods for introspection and counterfactual inference tasks that require causal relational reasoning among the causal, logic, and spatial-temporal structures of the visual and linguistic content. These results show that our CMQR has strong ability in modeling multi-level interaction and causal relations between the language and spatial-temporal structure.

To evaluate the generalization ability of  CMQR on other event-level datasets, we conduct experiments on TGIF-QA, MSVD-QA, and MSRVTT-QA datasets, as shown in Table \ref{Table5}-\ref{Table7}. From Table \ref{Table5}, we can see that our CMQR achieves the best performance for \emph{Action} and \emph{FrameQA} tasks. Additionally, our CMQR also achieves relatively high performance for \emph{Transition} and \emph{Count} tasks. For the \emph{Transition} task, the CMQR also outperforms nearly all comparison methods. For the \emph{Count} task, we also achieve a competitive MSE value. From Table \ref{Table6}, our CMQR outperforms all the comparison methods by a significant margin. For \emph{What}, \emph{Who}, and \emph{When} types, the CMQR outperforms all the comparison methods. It can be observed in Table \ref{Table7} that our CMQR performs better than the  best performing method IGV \cite{li2022invariant}. For \emph{What}, \emph{Who}, and \emph{When} question types,  the CMQR performs the best. Moreover, our CMQR can generalize well across different datasets and has good potential to model multi-level interaction and causal relations between the language and spatial-temporal structure.

\vspace{-10pt}
\subsection{Ablation Studies}
\vspace{-5pt}

We further conduct ablation experiments to verify the contributions of five essential components: 1) Explicit Causal Scene Learning (ECSL), 2) Visual Front-door Causal Intervention (VFCI), 3) Visual Causality Discovery (VCD), 4) Visual Causality Discovery (VCD), and Interactive Visual-Linguistic Transformer (IVLT). From Table \ref{Table8}, our CMQR achieves the best performance across all datasets and tasks. Without ECSL, the performance drops significantly due to the lack of the causal scene. This shows that the ECSL indeed find the causal scene that facilitate question reasoning. The performance of CMQR w/o ECSL, CMQR w/o VFCI are all lower than that of the CMQR. This validates that both the causal scene and visual front-door causal intervention are indispensable and contribute to discover the causal structures, and thus improve the model performance. The performance of CMQR w/o IVLT is higher than that of CMQR w/o VCD shows that visual and linguistic causal intervention modules contribute more than the IVLT due to the existence of cross-modal bias. With all components, our CMQR performs the best because all components contribute to our CMQR.

To validate the effectiveness of our causal module VCD in non-causal models, we apply the VCD to three state-of-the-art models Co-Mem \cite{gao2018motion}, HGA \cite{jiang2020reasoning} and HCRN \cite{le2020hierarchical}. As shown in Table \ref{Table11},  our VCD brings each backbone model a sharp gain across all benchmark datasets (+0.9\%$\thicksim$6.5\%), which evidences its model-agnostic property. To be noticed, for the causal and temporal questions (i.e., SUTD-TrafficQA), our VCD shows equivalent improvements on all four backbones (+1.05\%$\sim$2.02\%). These results validate that our VCD is effective in capturing the causality and reducing the spurious correlations across different models.

To conduct parameter analysis and validate the whether our CMQR could generalize to different visual appearance and motion features, we evaluate the performance of the CMQR using different visual backbones, please refer to Appendix 5. More visualization analysis are given in Appendix 6.

\begin{table}[!t]\renewcommand\tabcolsep{3pt}\renewcommand\arraystretch{0.9}\scriptsize
\begin{center}
\begin{tabular}{lccccc}
\hline
\multirow{2}*{Datasets}&CMQR w/o&CMQR w/o&CMQR w/o&CMQR w/o&\multirow{2}*{CMQR}\\
&ECSL&VFCI&VCD&IVLT&\\\hline
SUTD&37.68&37.42&37.28&37.75&\textbf{38.63}\\
MSVD&44.9&44.7&43.6&44.8&\textbf{46.4}\\
MSRVTT&38.1&38.0&37.5&37.7&\textbf{38.9}\\\hline
\end{tabular}
\end{center}
\vspace{-10pt}
\caption{Ablation study on three datasets. }
\vspace{-20pt}
\label{Table8}
\end{table}

\begin{table}[!t]\renewcommand\tabcolsep{2.0pt}\renewcommand\arraystretch{0.9}\scriptsize
\begin{center}
\begin{tabular}{lccc}
\hline
Models&SUTD-TrafficQA&MSVD-QA&MSRVTT-QA\\\hline
Co-Mem \cite{gao2018motion}&35.10&34.6&35.3\\
Co-Mem \cite{gao2018motion}+ VCD&\textbf{37.12} (+2.02)&\textbf{40.7} (+6.1)&\textbf{38.0} (+2.7)\\\hline
HGA \cite{jiang2020reasoning}&35.81&35.4&36.1\\
HGA \cite{jiang2020reasoning}+ VCD&\textbf{37.23} (+1.42)&\textbf{41.9} (+6.5)&\textbf{38.2} (+2.1)\\\hline
HCRN \cite{le2020hierarchical}&36.49&36.1&35.6\\
HCRN \cite{le2020hierarchical}+ VCD&\textbf{37.54} (+1.05)&\textbf{42.2} (+6.1)&\textbf{37.8} (+2.2)\\\hline
Our Backbone&37.42&44.7&38.0\\
Our Backbone + VCD&\textbf{38.63} (+1.21)&\textbf{46.4} (+1.9)&\textbf{38.9} (+0.9)\\\hline
\end{tabular}
\end{center}
\vspace{-10pt}
\caption{The VCD module is applied to non-causal models.}
\vspace{-32pt}
\label{Table11}
\end{table}

\vspace{-10pt}
\section{Conclusion}
\vspace{-5pt}
In this paper, we propose an event-level visual question reasoning framework named Cross-Modal Question Reasoning (CMQR), to explicitly  discover cross-modal causal structures. To explicitly discover visual causal structure, we propose a Visual Causality Discovery (VCD) architecture that learns to discover temporal question-critical scenes and mitigate the visual spurious correlations by front-door causal intervention. To align the fine-grained interactions between linguistic semantics and spatial-temporal visual concepts, we build an Interactive Visual-Linguistic Transformer (IVLT). Extensive experiments on four datasets well demonstrate the effectiveness of our CMQR for discovering visual causal structure and achieving robust event-level visual question reasoning.
%
%
%
 \bibliographystyle{splncs04}
%
\vspace{-20pt}
\bibliography{egbib}
\vspace{-20pt}
\end{document}